\documentclass[letterpaper, 10 pt, conference]{ieeeconf}  

\IEEEoverridecommandlockouts                              
\usepackage{graphicx}

\usepackage[utf8]{inputenc}

\usepackage[T1]{fontenc}
\usepackage{amsmath} 

\usepackage{arydshln}

\usepackage[numbers]{natbib}

\usepackage{todonotes}

\usepackage{pdfpages}

\usepackage{url}

\usepackage{mathtools}

\usepackage{bm}

\usepackage{subcaption}
\usepackage{ amssymb }
\usepackage{textgreek}

\usepackage{textcomp}

\usepackage{color,transparent}
\usepackage{amssymb}                            
\usepackage{graphicx}                           
\usepackage[pdfborder={0 0 0}]{hyperref}        
% ...
\usepackage{fixltx2e} 
\usepackage{chngpage}
\usepackage{float}

\usepackage{xcolor}

\title{\LARGE \bf
Recognizing and Tracking High-Level, Human-Meaningful Navigation Features of Occupancy Grid Maps}

\author{Payam Nikdel and Richard Vaughan$^{1}$\\
\thanks{$1$ Autonomy Lab, School of Computing Science, Simon Fraser University, Canada. {\tt\small \{pnikdel,vaughan\}@sfu.ca}}
}

\begin{document}
\maketitle

	\pagestyle{empty}
\begin{abstract}

	\par This paper describes a system whereby a robot detects and track human-meaningful navigational cues as it navigates in an indoor environment. It is intended as the sensor front-end for a mobile robot system that can communicate its navigational context with human users. From simulated LiDAR scan data we construct a set of 2D occupancy grid bitmaps, then hand-label these with human-scale navigational features such as closed doors, open corridors and intersections. We train a Convolutional Neural Network (CNN) to recognize these features on input bitmaps. In our demonstration system, these features are detected at every time step then passed to a tracking module that does frame-to-frame data association to improve detection accuracy and identify stable unique features. We evaluate the system in both simulation and the real world. We compare the performance of using input occupancy grids obtained directly from LiDAR data, or incrementally constructed with SLAM, and their combination. 
	
\end{abstract}

\section{Introduction}
Humans and robots can navigate by detecting and referring to distinctive features in their environment. When giving each other indoor navigation instructions, humans tend to refer to sparse structural features such as doors, corridors, staircases, etc. Most current robot navigation systems use very different features, such as occupancy grids in 2D metric space, corner features in image space, or nodes and edges in topological space. Recent work in semantic mapping allows robots to reliably identify features that are salient to humans. In this work we seek to bridge a gap in navigation representations between humans and robots, so that they can be shared in joint human-robot applications. Specifically, we train a CNN to take as input traditional occupancy grids and localize some structural features that humans use for navigation:  open doorways, closed doorways, and corridor openings at intersections. As a demonstration, we track these over time to obtain a sparse feature map that could be announced verbally to a visually-impaired user to assist them in navigation.

\par For a robot navigating in a known environment, it is possible to provide it with a map annotated with these semantic labels.  For example, using a prior annotated map containing the locations of hazards, doors and intersections, the robot can warn the user not to enter dangerous places while localizing itself in the environment. 

\par However, if the robot is located in an unknown environment, we can exploit machine learning methods to recognize different environmental features. For instance, \citeauthor{capi2012assisting} \cite{capi2012assisting} analyzed 2-dimensional (2D) LiDAR data using clustering technique to detect obstacles, steps and stairs, which makes it possible to warn the user about collisions or hazards. 


\par  In this study, we build a system to describe the navigational options ahead of the robot using occupancy grids obtained from either the 2D LiDAR data, SLAM or their combinations. To do so, we design a convolutional neural network (CNN) to detect open-rooms, closed-rooms and intersections around the robot in an unknown environment. A tracking module will then associate the model's predictions frame-to-frame to locate the target classes more precisely and obtain a sparse metric map of semantic features. Together, this system can describe the environmental navigational cues for the user (e.g. open-room on the left)
\par This approach can be used as a real-time detector of navigational options around a robot. It can also be used offline to annotate unlabelled occupancy grid maps. 

\par The contributions of this paper are:
\begin{itemize}
	\item A labelled dataset for training and evaluation;
	\item A system that is able to detect multiple navigational options around the robot in real-time using 2D LiDAR data, SLAM maps or their combination;
	\item A tracking module that does data association over time to maintain a sparse metric-semantic map;
	\item An experimental evaluation of the system in both simulation and the real world.
\end{itemize}

\begin{figure}[t] 
	\centering
	\includegraphics[width=0.94\columnwidth]{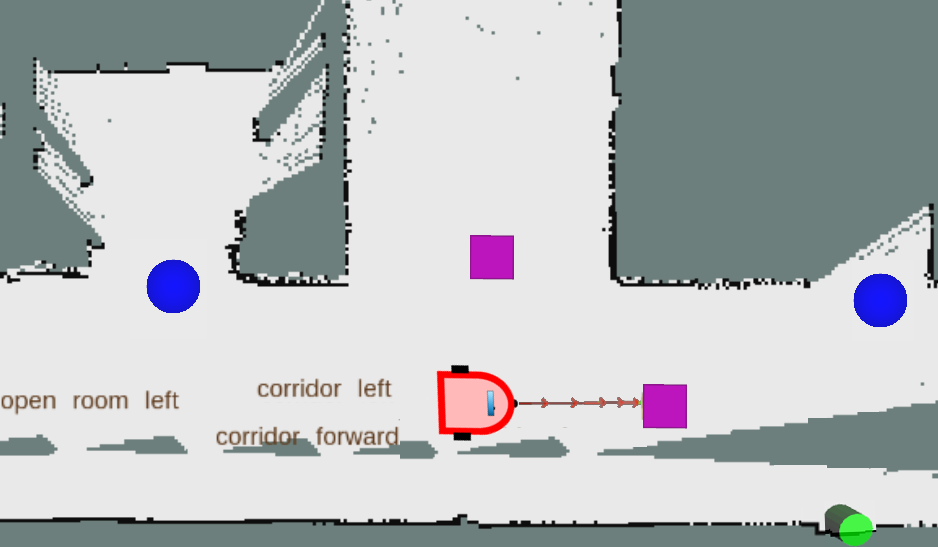}
	\caption{An example of a mobile robot describing the navigational options as it traverses through an indoor environment. Navigational options include the positions of open-rooms, closed-rooms or paths in an intersection.}
	\label{fig:cool image}
	\vspace{-7pt}
\end{figure}
\label{chapter:robotDescribe}

\section{Related Work}

\par There is a long history of robots and humans sharing spatial descriptions, going back to SRI's Shakey in the 1960s and 1970s \cite{shakey}. There is a large body of work on robots that can understand natural human commands \cite{tellex2011understanding, yi2016expressing,zheng2017navigation}, while fewer studies have focused on robots that can translate their observations to descriptions or instructions. In work explicitly considering human-robot interaction systems, Skubic et al. \cite{skubic2004spatial} investigated spatial semantic models for human-robot dialogue where the robot describes the spatial relation of objects with respect to itself. Daniele \citeauthor{daniele2016navigational}  proposed a navigational guide system able to generate  instructions for navigating from point $A$ to point $B$ given a known map using a sequence-to-sequence Recurrent Neural Network (RNN).

\par In the context of multi-agent systems where agents collaborate towards a goal, communication plays a significant role. 
\citeauthor{andreas2017translating} \cite{andreas2017translating} formulated the problem of interpreting the policy of agents as translating their messages to human language. They build a translation model upon a similarity criteria to facilitate interpretation of communications between collaborative agents.

\par Various studies have used 2D \cite{capi2012assisting, goeddel2016learning, pronobis2017learning} and 3D \cite{Guan2015tree,Guan2016review} LiDAR scanners for object detection. Among these studies, 3D LiDAR scanners on mobile vehicles are able to collect more accurate and efficient 3D information about the surrounding environment. This type of mobile LiDAR scanning has gained popularity in  road mapping studies. \citeauthor{Guan2016review} \cite{Guan2016review} surveyed recent studies that use LRF data to detect and extract road surfaces, on-road structures and pole-like objects. 3D LiDAR classification is also being used in the ecological analysis. For instance, \citeauthor{Guan2015tree} \cite{Guan2015tree} proposed a tree classification method to classify different species of trees using mobile LiDAR data. 

\par 2D LiDAR data is often used to classify locations (e.g. corridors,  doorways or different rooms) \cite{goeddel2016learning, pronobis2017learning} or detect big environmental objects (e.g. stairs, steps or obstacles) \cite{capi2012assisting}.  \citeauthor{goeddel2016learning} \cite{goeddel2016learning} used CNN models to classify objects in the environment, taking 2D LiDAR data as the model's input. Likewise, in another study, \citeauthor{pronobis2017learning} \cite{pronobis2017learning} propose a probabilistic framework using Sum-Product Networks (SPNs) and deep learning to classify locations in an autonomous robot application. Although these two studies are similar to the system we present in this paper, they only produce one target class per 2D LiDAR frame. By contrast, we detect multiple target classes from each data frame (occupancy grid obtained from either 2D LiDAR, SLAM or their combinations) and we localize them with sub-pixel accuracy on the map, i.e. in navigation space.  A tracking module then aggregates the frame-to-frame predictions to improve the accuracy and position estimates of detected target classes.

\section{Approach}
Given a sequence of a robot's sensor data (2D LiDAR data, IMU and odometry), we create and annotate a map of the environment while describing the navigational options for the user. Our system can be used in a robotic system designed to guide a user in an unknown indoor environment. This system could provide feedback about the location of closed-rooms, open-rooms and corridors in an intersection. For instance, the robot can interact with the user by uttering sentences such as ``Open-room on the right'' or  ``corridors on the left and right'' (for a three-way intersection with two navigational options excluding the corridor occupied by the robot). To achieve this, as the robot navigates through indoor maps, it should detect and track the positions of target classes as it navigates through the environment.

\subsection{System Overview}
Our system is implemented in the Robot Operating System \cite{quigley2009ros} (ROS)  and tested in simulation in Stage \cite{vaughan:stage08} and in the real world on a Clearpath Husky robot.

\par We investigate three variations  of local occupancy grid maps, one obtained from a single LiDAR scan, one obtained by metric LiDAR/inertial SLAM, and their combination. The laser scans are obtained from a Sick LMS-111 LiDAR with 270\textdegree{} field of view. Laser scans are rendered into a 2D occupancy grid local-map, and we use the ROS implementation of grid mapping (GMapping) for SLAM \cite{grisetti2007gmapping} to build an occupancy grid map of the environment. We will refer to this map as the "GMap". This occupancy grid map is created incrementally using the odometry data, the inertial measurement unit (IMU) data and the 2D LiDAR data. By cropping the 16-by-16 meter local-map in front of the robot, we obtained the GMap local-map. 
\par To navigate in the environment, the robot uses the ROS navigation stack\footnote{\url{http://ros.org/wiki/navigation}} along with a trajectory planning module based on Dynamic Window Approach \cite{580977}.

\par For our network model, we use a combination of convolutional, Batch Normalization and ReLU layers with residual connections (see Section \ref{subsection:Architecture} for more details). The inputs to our model are the laser local-map and the GMap local-map.

On top of this network, our system uses a tracking module that aggregates predictions from individual frames and improves the location accuracy of target classes. The tracked predictions are then passed to our describe module that narrates the navigational options to the user.

\subsection{Dataset}

\begin{figure}[h]
	\centering
	\includegraphics[width=.9\columnwidth]{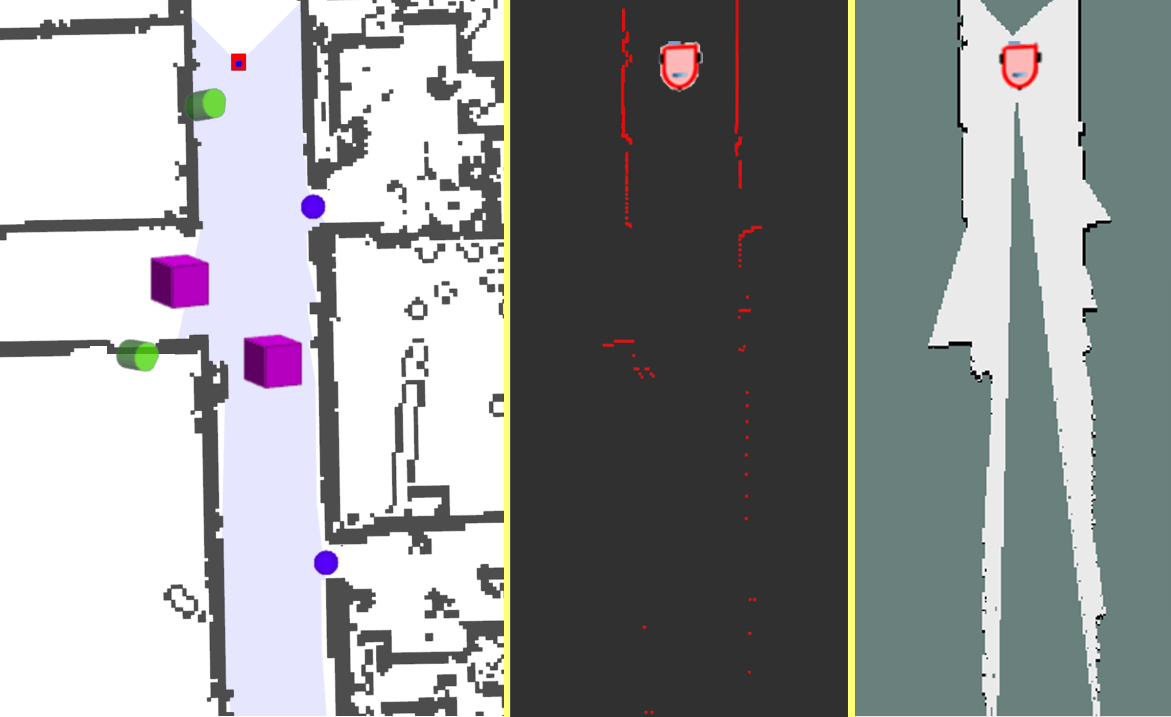}
	\caption{ Visualization of one frame of data during dataset generation. Left:  visualization from the Stage simulator showing a robot (top) in a corridor of a map obtained from a real building, overlaid with ground-truth annotations of closed-rooms (green cylinders), open-rooms (blue spheres) and corridor intersections (purple cubes). Centre: corresponding 2D LiDAR data in RViz. Right: local occupancy map constructed by GMap SLAM.}
	\label{fig:dataset}
	\vspace{-7pt}
\end{figure}

\label{subsection:dataset}
We generate our own dataset containing the raw laser data (an array containing the angle of each hit and its distance from the LRF), the current local-map (a 16-by-16 meter map in front of the robot) from the GMap, and positions and labels of the target classes (navigational objects). To gather the data, we used the Stage robot simulator \cite{vaughan:stage08} loaded with real world occupancy grid maps. We include six preexisting occupancy grid maps \cite{Radish, mozos2005supervised} (with small modifications) and two new ones that were created by performing SLAM in the Applied Science Building at Simon Fraser University. Figure \ref{fig:dataset} shows one frame of the dataset generation process using the Stage simulator. 

\par For each map, we annotate the open-rooms, closed-rooms and the beginning of each corridor at all intersections. The total number of target labels in these eight maps is shown in Table \ref{tab:annotationmaps}. To generate our dataset, we divide the maps into training and testing portions. Two of our maps are used only for training, one map only for testing, and the other five are divided into two parts. We then move the robot in predetermined paths in each map, while recording the coordinates of the robot  until it explores each corridor in that map at least twice. These consecutive recorded coordinates form trajectories that are used to generate data. The total number of labels per target classes is shown in Table \ref{tab:dataset_annotation}.

This small amount of training data was extensively augmented to be sufficient for training. One unusual aspect is that we synthesized partially-open doors at various angles in doorways, since we required the model to robustly identify doorways irrespective of door angle despite large difference in appearance in a 2D top down view. 

\begin{table}
	\caption{Number of labels per target classes in the train, validation and test portion of the dataset.}
	\centering
	
	\begin{tabular}{rrrrr}
		
		& Closed Room          & Open Room           & Corridor              & Total       \\
		& \%          & \%            & \%             &        \\ \cline{2-5} 
		\multicolumn{1}{l:}{Train} & 29                 & 46                & 25                 & 33316                 \\
		\multicolumn{1}{l:}{Validation} &29	            &46	                &25	                &6663                \\
		
		\multicolumn{1}{l:}{Test}  & 28                 & 39                 & 33                 & 15610                 
	\end{tabular}
	\label{tab:dataset_annotation}
\end{table}


\newcommand{\specialcell}[2][c]{%
	\begin{tabular}[#1]{@{}c@{}}#2\end{tabular}}

\begin{table}[h]
	\caption{Total number of annotations for each one of our target classes.}
	\centering
	\begin{tabular}{cccc}
		Closed Room & Open Room & Corridor & Total\\ \hline
		232         & 121       & 95 & 448
	\end{tabular}
	\label{tab:annotationmaps}
\end{table}

\begin{figure*}[h]
	\centering
	\includegraphics[width=\textwidth]{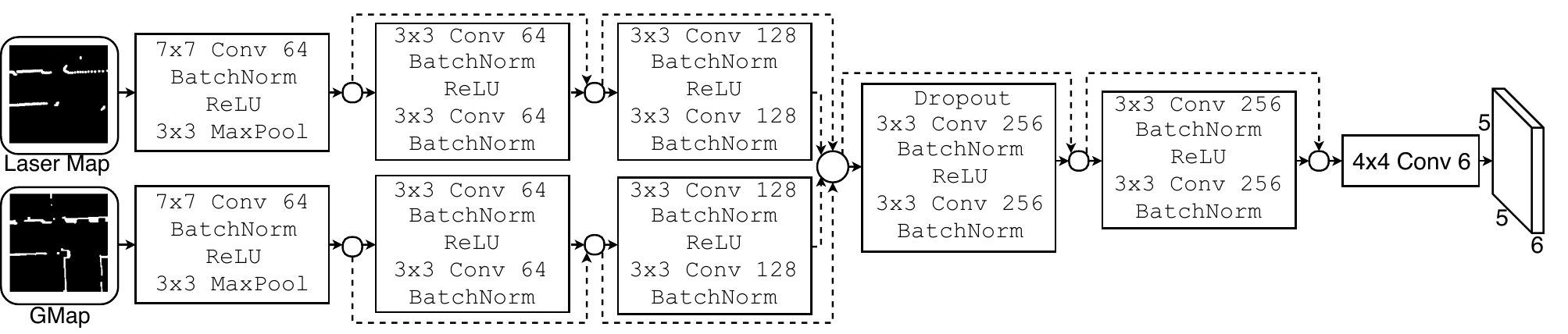}
	\caption{Our network architecture is a fully convolutional model with residual connections. The inputs are two local-map images, $244  \times 244$, one generated from 2D LiDAR and the other one from the GMap during run-time.  The output is a $5 \times 5  \times 6$ tensor, which is the grid representation of predictions. Each grid cell contains the confidence score, coordinates and probability of target classes for that grid cell.}
	\label{fig:architecture}
	\vspace{-8pt}
\end{figure*}

\subsection{Data Augmentation}
\label{subsection:dataaugmentation}

To make the model more robust to unseen data, we apply data augmentation that includes rotating, translating and re-sizing the local-map images (both the laser and the GMap local-maps). We also add different orientations of the door for the open-room class.

\subsubsection{Rotation, Translation or Re-size}
\begin{figure}[h]
	\centering
	\includegraphics[width=\columnwidth]{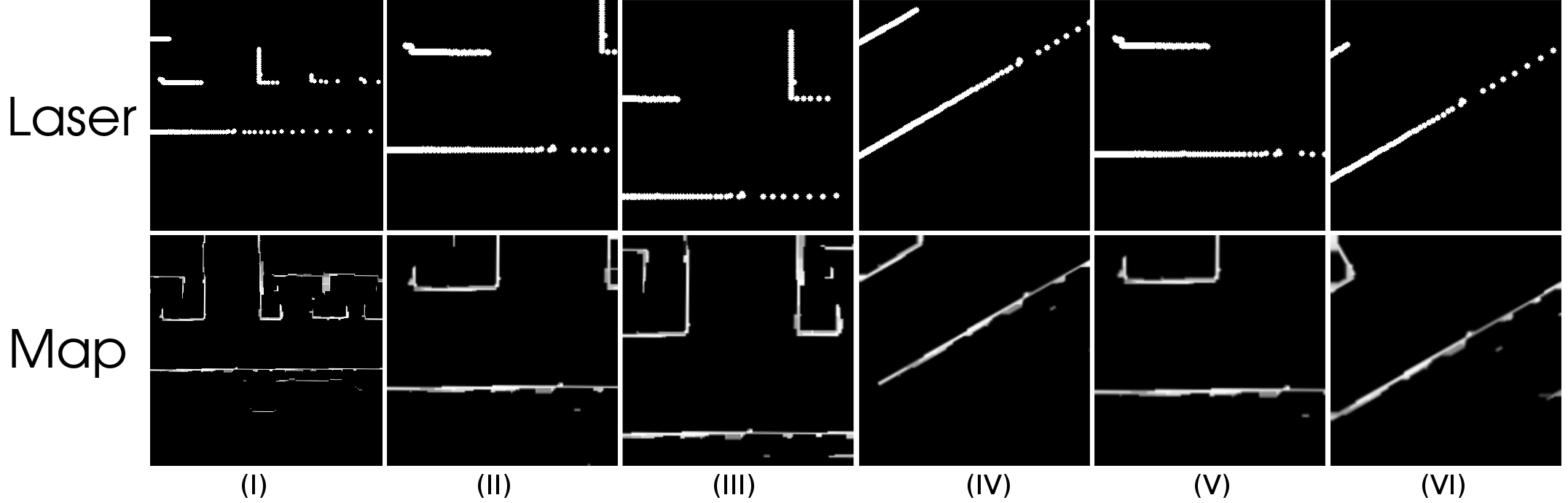}
	\caption{Data augmentation process. Top: the laser local-maps, Bottom: the GMap local-maps. (I) the original 16-by-16 meter local-maps, (II) the cropped 8-by-8 meter local-maps centered at the robot, (III) translation of +1.6 meter in both x and y directions, (IV) rotation by -30\textdegree{}, (V) resize by 1.2 times and (VI) the obtained augmentation by applying these operation in order. (III) to (VI) are cropped 8-by-8 meter local-maps.}

	\label{fig:data_augmentation}
	
\end{figure}

During training, whenever we fetch a GMap local-map or 2D LiDAR data from the dataset, we randomly rotate, translate or re-size them. Figure \ref{fig:data_augmentation} shows the possible data augmentation operations. The resulting images are fed into our neural network. 

The primary GMap local-map is saved with dimensions twice as big as the images the network needs. It helps to do the data augmentation by preventing unknown pixels in the final map. After the operations, we crop the GMap local-map into an 8-by-8 meter map and resize it to the desired network input size of $244\times244$ pixel. The same operations (rotation, translation and re-size) are also applied to the laser scan data by first converting the 1D laser array to 2D local-map coordinates.

\subsubsection{Adding New Doors' Orientations}
\begin{figure}[h]
	\centering
	\includegraphics[width=\columnwidth]{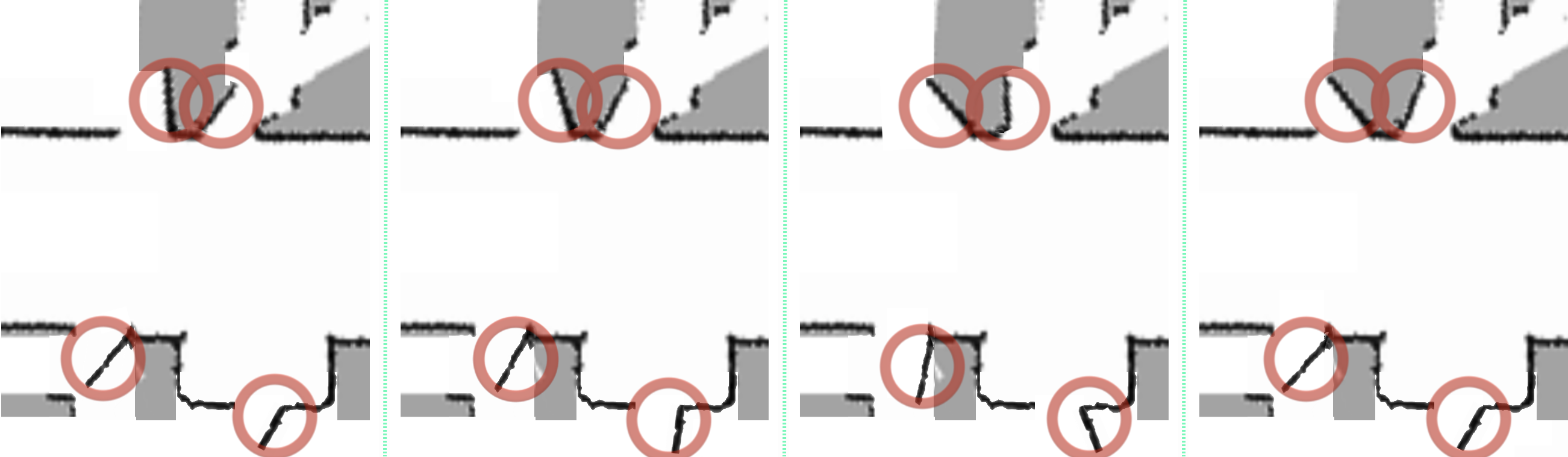}
	\caption{Four different orientations of four doors.  Doors are annotated by red circles. During data augmentation, we add different orientations for the door panels to avoid over-fitting. We used a random angle between 30 and 100 degrees for open doors.   }
	\label{fig:door_orrientation}
	
\end{figure}
\label{subsubsection:door_orrientation}
In this step, we create new maps by modifying the previous maps. First, we annotate the position and width of each door panel in open-rooms; then we remove the door panel from the occupancy grid maps. Next, we generate new maps by randomly adding open-rooms. For open-rooms, we assume that the angle between the door frame and door panel is between 30\textdegree{} and 100\textdegree{}. An example of four different doors' orientations is illustrated in Figure \ref{fig:door_orrientation}. 

\begin{figure*}[h]
	\centering
	\includegraphics[width=\textwidth]{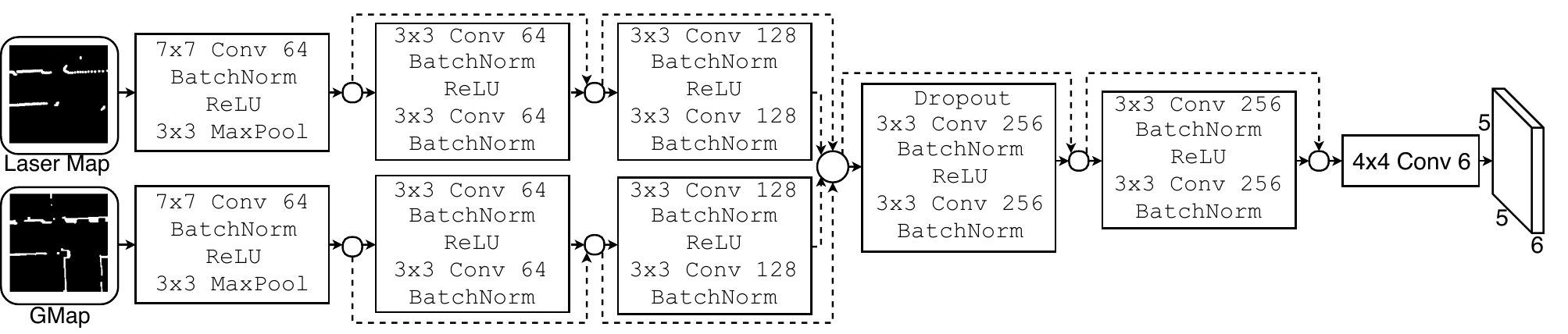}
	\caption{Our network architecture is a fully convolutional model with residual connections. The inputs are two local-map images, $244  \times 244$, one generated from 2D LiDAR and the other one from the GMap during run-time.  The output is a $5 \times 5  \times 6$ tensor, which is the grid representation of predictions. Each grid cell contains the confidence score, coordinates and probability of target classes for that grid cell.}
	\label{fig:architecture}
	\vspace{-8pt}
\end{figure*}

\subsection{Architecture}
\label{subsection:Architecture}

\par We trained three different models to predict the position of the target classes around the robot. All of these models use a similar network architecture but with different input data. We define these three models as follows:
\begin{itemize}
	\item \textbf{Laser model}, which uses only the laser local-map
	\item \textbf{Map model}, which uses only the GMap local-map
	\item \textbf{Combined model}, which uses both GMap local-map and laser local-map
\end{itemize}
\par The network architecture is presented in Figure \ref{fig:architecture}. The first part of the network consists of two parallel networks where one extracts features from the GMap local-map and the other from the laser local-map. Each part of the parallel network consists of three ResNet blocks with residual connections. Each ResNet block contains convolutional, batch normalization and ReLU layers.  In the next stage, the concatenation of these two parallel networks forms the input of the next two ResNet blocks followed by a $4 \times 4$ convolutional layer.

\subsection{Implementation details}
\label{subsection:implementation}

\par Our architecture is inspired by ResNet34 \cite{he2016deep} and YOLO9000 \cite{redmon2016yolo9000}. The inputs of our network are two $244 \times 244$ pixel images of the surrounding environment, belonging to a laser local-map or a GMap local-map. Inspired by YOLO's methodology, we divide each input image into a 5 by 5 grid (referred to as $g$). For each grid cell, we define several properties including confidence score, coordinates and the probability of target classes. Therefore, the model predicts a $5 \times 5 \times 6$ tensor, where the numbers along the 3rd axis correspond to predicted properties of each grid cell.  Each vector $v_{i,j,*}$ is responsible for detecting the object that resides in the grid cell of $g(i,j)$ with six properties: 
Confidence score, $(x,y)$ coordinates and conditional class probabilities for the three target classes.  The confidence score is the estimate of model certainty about whether the center of an object resides in this grid cell. The $(x,y)$ coordinates are the positions of each object's center relative to the top-left corner of the grid cell in the input image ($0< x,y < 1$). The last three conditional class probabilities $Pr(Class_i|Object), i=0,1,2$ are the probabilities of each target class if an object of class $i$ exists in this grid cell. In our case, object classes are open-room, closed-room and the beginning of corridors. To calculate the final probability of each target class, we multiply the conditional class probabilities of each target class by the cell confidence score:
\begin{equation}
Pr(Class_i) = Pr(Class_i|Object)*Pr(Object)
\label{eq:probability_classes}
\end{equation}

\par To train our network to predict the $5 \times 5 \times 6$ tensor, we use the combination of Cross-Entropy loss and L2 loss (mean squared error). The Cross-Entropy loss  ($X_{ent}$) is used to learn the conditional class probabilities and the L2 loss used to learn both the coordinates position ($\ell ^p _2$) and confidence score  ($\ell ^c _2$) of each target classes. Our final loss function is a weighted sum of these losses:
\begin{equation}
\mathcal{L} = \alpha_1 * X_{ent} + \alpha_2 * \ell ^p_2 + \alpha_3 * \ell ^c _2
\end{equation}
\par where, 
\begin{equation}
\alpha_1 + \alpha_2 + \alpha_3 = 1
\end{equation}

\par In these equations, $\alpha_1, \alpha_2$ and $\alpha_3$ are weights of the loss functions. We set $\alpha_1=0.61$, $\alpha_2=0.14$ and $\alpha_3=0.25$, although we tested other values empirically and observed minor variation in performance. All three models are trained using the Adadelta optimizer \cite{zeiler2012adadelta} with a batch size of 60 and randomly initialized weights on the training dataset. During training, we evaluate the model at each epoch on the validation dataset and keep the weights of the model that has the minimum total loss. These models were implemented with the PyTorch library \cite{paszke2017automatic} and were trained on an Nvidia GeForce GTX 1080 Ti GPU.

\subsection{Tracking Module}
\label{subsection:tracking_system}

\begin{figure}[h]
	\centering
	\includegraphics[width=0.75\columnwidth]{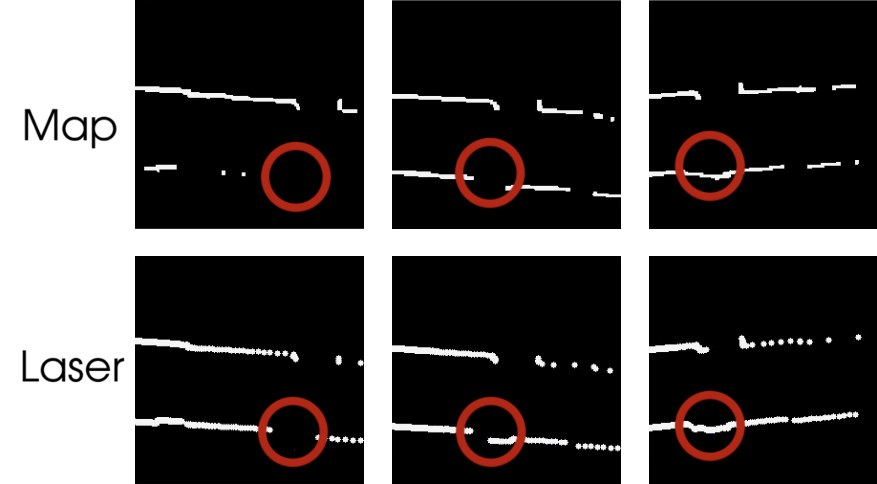}
	\caption{An example of local-maps when the robot approaches a closed-room. The closed-room is annotated by a red circle. The closed-room becomes clearer as the robot comes closer to the closed-room. In the beginning, the closed-room looks like an open-room but later (on the second frame for the laser local-map and the third frame for the GMap local-map) it can be distinguished as a closed-room.}
	\label{fig:closetoopen}
	\vspace{-8pt}
\end{figure}

In our application, a robot should be able to describe the navigational options for the user at the appropriate moments. Consequently, our system should be able to track the surrounding objects as the robot approaches them, and the robot should narrate the labels at the appropriate time. Our tracking module ensures that our system reliably tracks detected objects between frames as the local map changes.

When the robot is moving along a path and approaches a room or an intersection, the target object may be in the field of view of the robot for a few frames. In each frame, the robot sees the object with a slightly different shape. The reasons for these differences are the movement of the robot, error in the robot's sensors and the incomplete map created by SLAM, which is being updated incrementally as the robot gets closer to the object.
For example, a closed-room in some cases (especially at a distance) might look like an open-room (see Figure \ref{fig:closetoopen}). Open-room and corridor also look similar in many cases. Thus, the network may change its prediction when the robot approaches an object. Moreover, when the shape of the object changes, the network finds slightly different coordinates for the labels. In other words, the system should not change the coordinates or class of an object based on a single network prediction. Instead, it should consider all the prediction labels and positions for that object.

Our tracking module works by first saving and calculating the position of the target labels on the GMap global map (conversion between different coordinate frames is done by the ROS TF package\footnote{http://wiki.ros.org/tf}), and then  clustering the nearest target points using the k-Nearest Neighbors algorithm (with a maximum distance of one meter). Next, for every target class in each cluster, we average over the final probability (obtained from Equation \ref{eq:probability_classes})  and assign the most probable target to the cluster. Finally, the coordinates of each cluster are calculated by averaging the coordinates of the most probable target class for that cluster.

\par The tracking module relies on the global coordinate frame that is updated by the GMapping algorithm. Although the SLAM algorithm usually builds a reliable map, sometimes it shifts the global coordinate frame or part of the occupancy grid maps. The reason for this is the cumulative errors in the odometry of the robot or delays in the sensor's data. These shifts add cumulative errors to the previous predictions of our tracking module. To reduce these cumulative errors, each point in our tracking module has a lifetime of 30 seconds after its detection.

\par Ultimately, using the most probable target label and position, the describe module explains the environment for the user with the following procedure. First, it finds the coordinates of the detected targets around the robot's position within five meters. Next, it calculates the angle $Q$  between the robot-object line and the robot orientation. At last, the describe module says the target class name followed by its position around the robot (e.g. closed-room left). The position of the object with respect to the robot is defined as follows:
\begin{itemize}
	\item left side, if -120\textdegree{}$<Q<$-50\textdegree{}
	\item front, if -30\textdegree{}$<Q<$-30\textdegree{}
	\item right side, if 50\textdegree{}$<Q<$120\textdegree{}
\end{itemize}

\section{Results and Discussion}

We evaluate our methodology by conducting three experiments. The results of each experiment are compared using three models described in Section \ref{subsection:Architecture}. These models are trained on the training dataset (see Section \ref{subsection:dataset}), and the best performing model is obtained by evaluating on the validation dataset. We tested the same trained model for all three experiments. On each experiment, we report the recall, precision and F1 score, which are three standard metrics for classification.

\subsection{Experiment One}

\begin{figure}[h]
	\centering
	\includegraphics[width=\columnwidth]{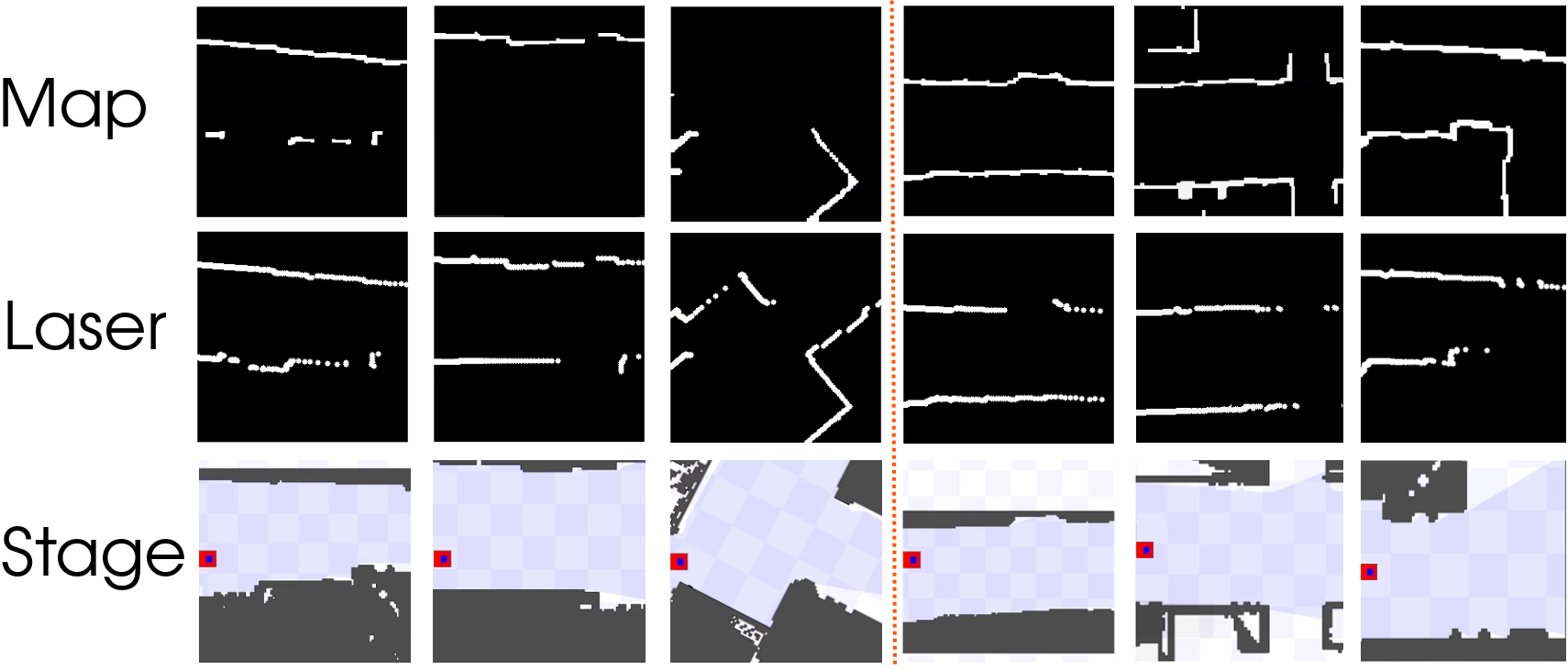}
	\caption{Comparison between 2D LiDAR local-maps and GMap local-maps. The top row shows the GMap local-map; the middle row shows the laser local-maps and the bottom row is the screenshot of the robot in the Stage simulator. On the left side, there are three cases in which laser local-maps are more detailed; this usually happens in an unexplored map when the robot changes its orientation. The right side shows three cases where the GMap data is more detailed. This often occurs when the robot is in a previously explored place.}
	\label{fig:laser_map}
	\vspace{-7pt}
\end{figure}

\begin{table*}
	\caption{Recall, precision and F1 score of three models running on the \textbf{validation} and \textbf{testing dataset}. The numbers indicate difficulties of detecting corridors (at the beginning of each intersection) compared to closed-rooms and open-rooms.}
	
	\begin{minipage}{0.5\textwidth}
		\centering
		\captionsetup[subtable]{labelformat=empty}
		\subcaption{Validation}
		\vspace{-5 pt}
		\subcaption{Recall}
		\begin{tabular}{ccccc}
			& Closed Room              & Open Room             & Corridor                \\ \cline{2-4} 
			\multicolumn{1}{c:}{Laser}& 0.84         & 0.91          & 0.77          \\
			\multicolumn{1}{c:}{Map}  & 0.82         & 0.89          & 0.77          \\
			\multicolumn{1}{l:}{Combined}  & \textbf{0.9} & \textbf{0.95} & \textbf{0.89}   
		\end{tabular}
		
		\subcaption{Precision}
		\begin{tabular}{ccccc}
			& Closed Room & Open Room & Corridor \\ \cline{2-4} 
			\multicolumn{1}{c:}{Laser} & 0.83          & \textbf{0.99} & 0.98          \\
			\multicolumn{1}{c:}{Map}  & 0.74          & \textbf{0.99} & 0.97          \\
			\multicolumn{1}{l:}{Combined} & \textbf{0.84} & 0.98          & \textbf{0.99}
		\end{tabular}
		
		\subcaption{F1 Score}
		\begin{tabular}{ccccc}
			& Closed Room & Open Room & Corridor \\ \cline{2-4} 
			\multicolumn{1}{c:}{Laser} & 0.84          & 0.95          & 0.86          \\
			\multicolumn{1}{c:}{Map}  & 0.78          & 0.94          & 0.86          \\
			\multicolumn{1}{l:}{Combined}  & \textbf{0.87} & \textbf{0.96} & \textbf{0.94}
		\end{tabular}
		
	\end{minipage}
	\begin{minipage}{0.5\textwidth}
		\centering
		\captionsetup[subtable]{labelformat=empty}
		\subcaption{Test}
		\vspace{-2 pt}
		\subcaption{Recall}
		
		\begin{tabular}{ccccc}
			& Closed Room              & Open Room             & Corridor                \\ \cline{2-4} 
			\multicolumn{1}{c:}{Laser} & 0.72          & 0.85          & 0.47          \\
			\multicolumn{1}{c:}{Map}  & 0.68          & 0.68          & 0.46          \\
			\multicolumn{1}{l:}{Combined}  & \textbf{0.78} & \textbf{0.86} & \textbf{0.48}
		\end{tabular}
		
		\subcaption{Precision}
		\begin{tabular}{ccccc}
			& Closed Room & Open Room & Corridor \\ \cline{2-4} 
			\multicolumn{1}{c:}{Laser} & \textbf{0.82} & \textbf{0.99} & 0.96          \\
			\multicolumn{1}{c:}{Map}  & 0.65          & 0.98          & 0.95          \\
			\multicolumn{1}{l:}{Combined} & 0.80          & \textbf{0.99} & \textbf{0.97}
		\end{tabular}
		
		\subcaption{F1 Score}
		\begin{tabular}{ccccc}
			& Closed Room & Open Room & Corridor \\ \cline{2-4} 
			\multicolumn{1}{c:}{Laser} & 0.77          & 0.91          & 0.63          \\
			\multicolumn{1}{c:}{Map}  & 0.67          & 0.80          & 0.62          \\
			\multicolumn{1}{l:}{Combined}  & \textbf{0.79} & \textbf{0.92} & \textbf{0.64}   
		\end{tabular}
		
	\end{minipage}
	\label{table:acc_recall_dataset}
	\vspace{-7pt}
\end{table*}

The First Experiment is conducted to evaluate the accuracy of our network without the use of the tracking module. We run each of the three models on the testing dataset.
Table \ref{table:acc_recall_dataset} shows the recall, precision and F1 score of each class for all three models in the validation and the testing datasets. These results show that the system obtains the highest recall for open rooms, and the lowest for corridors. Open-rooms are probably easier to detect for several reasons. First, we train our network with different orientations of doors (for details see Section \ref{subsection:dataaugmentation}), which makes it more robust to open-rooms. Secondly, door frames usually have similar dimensions between different buildings. Lastly, in the case that the network has learned the relation between free spaces,  it may easily detect an open-room as a narrow space that connects two wide open-areas as opposed to a closed-room, which is a small recess inside the wall. Another reason behind the poor recall of the corridors is likely the variety of intersection types. For example, the size of intersection corridors can vary widely. In some cases, the three-way intersections are very similar to a corridor with an open room on the side.
All the models are less precise at detecting closed-rooms compared to other classes. It might be due to false identification of some of the recesses in the wall as a closed room, resulting in lower precision.

\par  We also observe that the F1 score of the Map model is lower than the other two models. We find that the GMap local-maps have fewer details compared to local-map from LiDAR data. Besides, when the robot is turning into an unknown area of the map, the occupancy grid map will only be updated a few frames later, because SLAM requires multiple measurements for each one of the obstacles in the LiDAR view. Although the GMap local-map may have fewer details compared to the laser local-map, a combination of them in the Combined model slightly enhances the F1 score in most cases.

\par We measure the run-time speed of our network as it is essential for a robotic application to have a fast network that can predict in real-time. Our network can run as fast as 140 Hz (around seven milliseconds for each run) with a system equipped with a GeForce GTX 1080 Ti or 12 HZ (around 83 milliseconds for each run) using an Intel i7-7700 CPU.

\subsection{Experiment Two}

\begin{figure}
	\centering
	\hfill
	
	\begin{subfigure}{1\columnwidth}
		\includegraphics[width=\columnwidth]{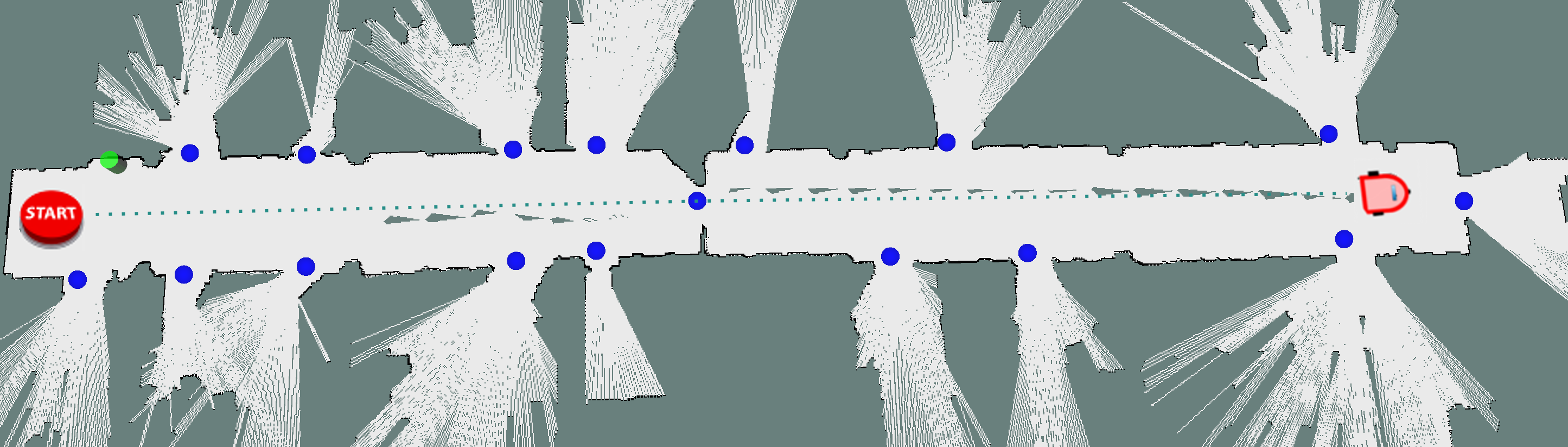}
		\subcaption[]{FR79 building (\textbf{train} and \textbf{validation})}
	\end{subfigure}
	\hfill
	
	\begin{subfigure}{1\columnwidth}
		\includegraphics[width=\columnwidth]{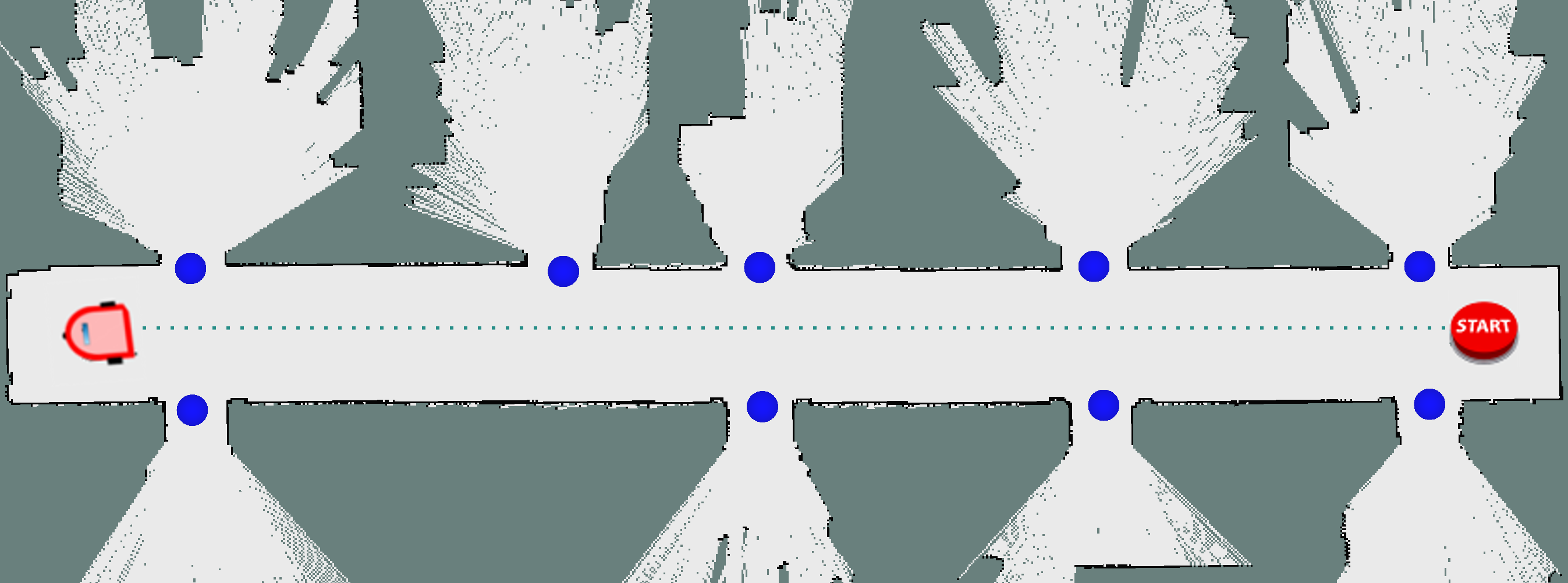}
		\subcaption[]{FR52 building (\textbf{test})}
	\end{subfigure}
	\hfill
	
	\caption{Prediction of our system for the FR52 and FR72 building map. Closed-rooms predictions are annotated by green cylinders and open-rooms by blue spheres.}
	\label{fig:fr}
	\vspace{-9pt}
\end{figure}

\begin{figure}
	\centering
	\hfill
	
	\begin{subfigure}{1\columnwidth}
		\includegraphics[width=\textwidth]{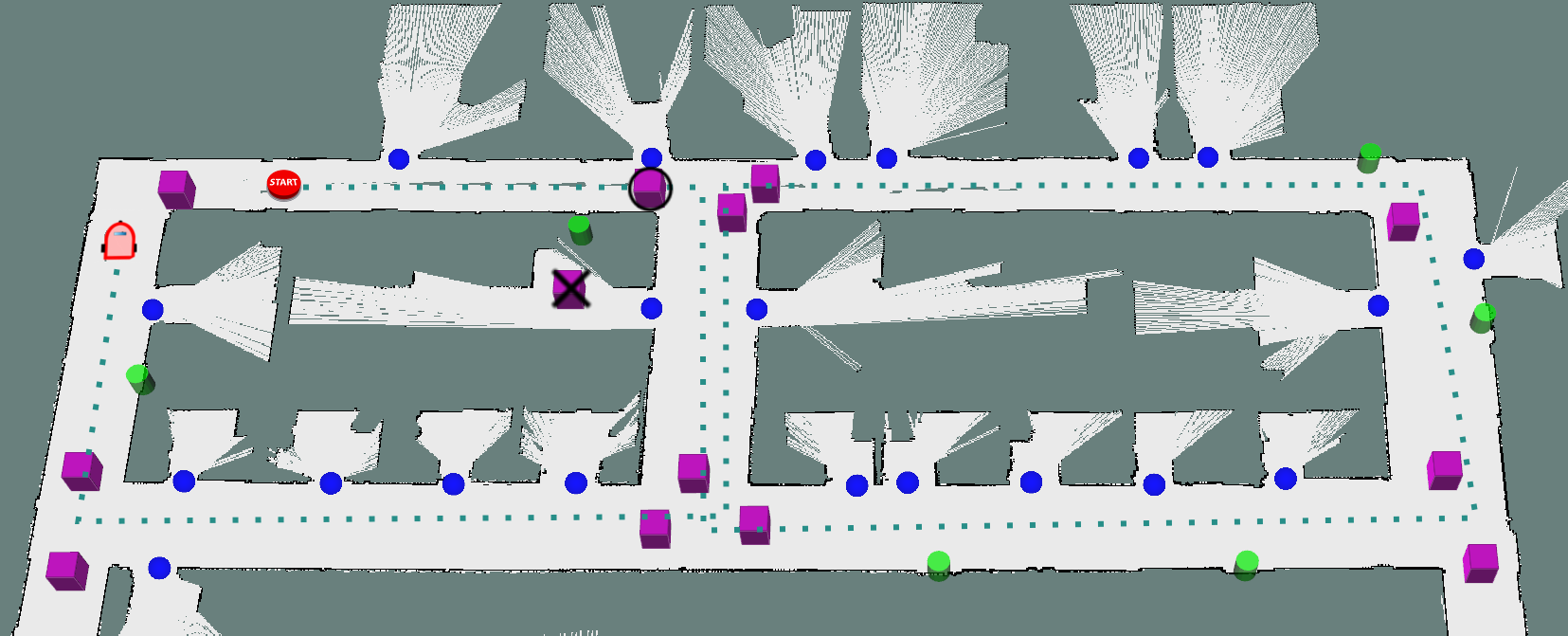}
		\subcaption[]{SAIC building (\textbf{train}  and \textbf{validation} portion)}
	\end{subfigure}
	\hfill
	
	\begin{subfigure}{1\columnwidth}
		\includegraphics[width=\textwidth]{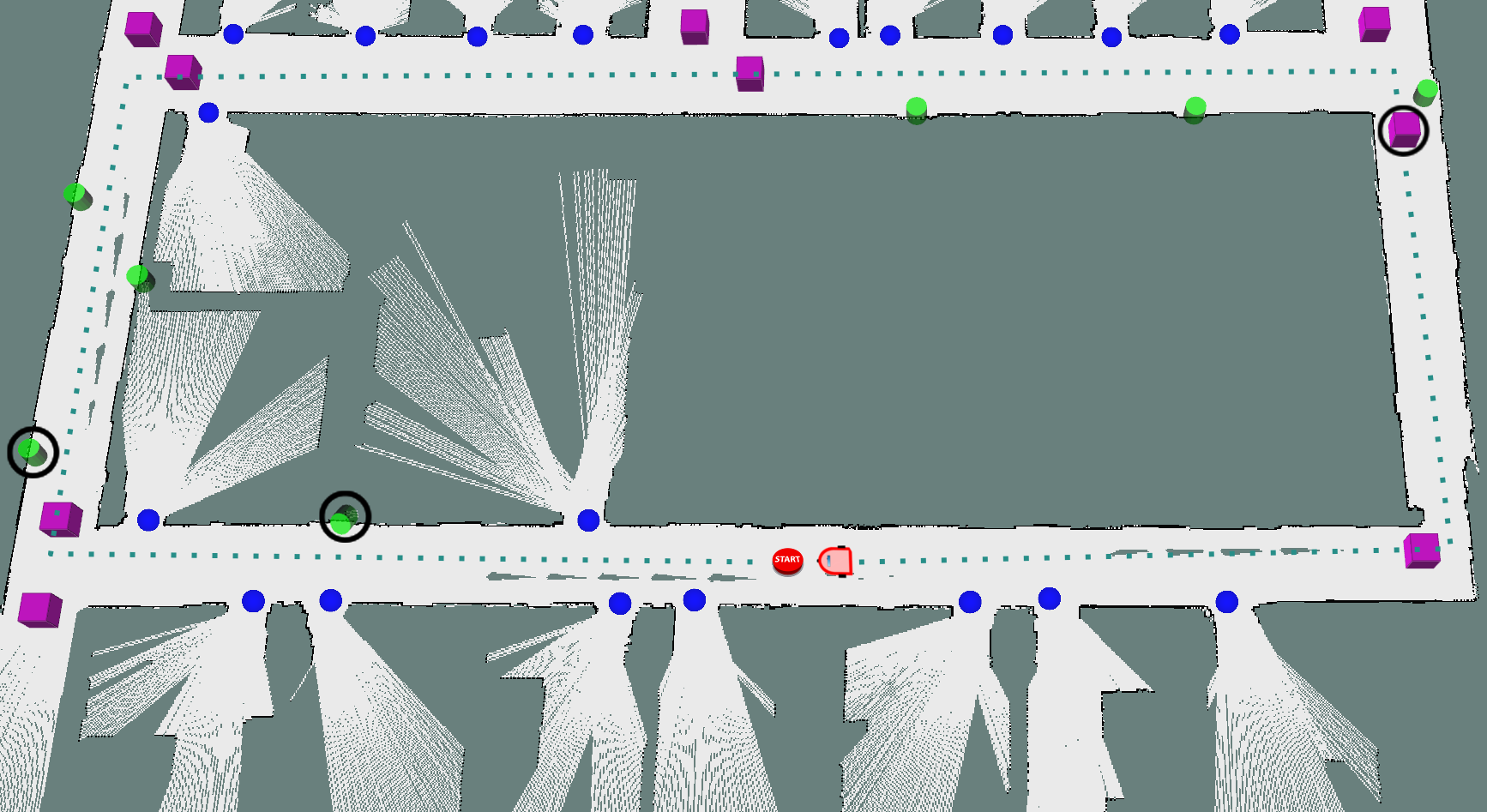}
		\subcaption[]{SAIC building (\textbf{test} portion)}
	\end{subfigure}
	\hfill
	
	\caption{Prediction of our system for the SAIC building map.  Closed-rooms predictions are annotated by green cylinders, open-rooms by blue spheres and the start of each corridor in the intersections by purple cubes. The false-positives are annotated by $\times$ and false-negatives by $\circ$ over predictions. }
	\label{fig:SAIC}
	\vspace{-9pt}
\end{figure}

In Experiment Two, we evaluate the addition of our tracking module on top of the model predictions. We also investigate the performance of our system in an unexplored environment versus a previously explored environment. We annotate new trajectories in our occupancy grid maps. These trajectories are chosen in a way to visit the whole map with minimal crossings (See Figures \ref{fig:fr} and \ref{fig:SAIC}). As a result, the robot would see most of the map for the first time.

\par  In the unexplored experiment, the robot follows a trajectory similar to the data generation phase. As the robot moves, SLAM GMapping updates the global occupancy grid map. In the explored experiment, the robot follows the same trajectory for the second time using an already explored map. 

\begin{table*}
	\caption{Recall, precision and F1 Score for all three models for \textbf{Experiment Two} in the \textbf{unexplored} and the \textbf{explored maps} using test occupancy grid maps.}
	
	\begin{minipage}{0.5\textwidth}
		
		\centering
		\captionsetup[subtable]{labelformat=empty}
		\subcaption{Unexplored Maps}
		\vspace{-5 pt}
		\subcaption{Recall}
		\begin{tabular}{ccccc}
			& Closed Room              & Open Room             & Corridor                \\ \cline{2-4} 
			\multicolumn{1}{c:}{Laser} & 0.75          & 0.83          & \textbf{0.74} \\
			\multicolumn{1}{c:}{Map}  & 0.63          & 0.79          & 0.61          \\
			\multicolumn{1}{l:}{Combined}  & \textbf{0.79} & \textbf{0.84} & 0.69         
		\end{tabular}
		
		\subcaption{Precision}
		\begin{tabular}{ccccc}
			& Closed Room & Open Room & Corridor \\ \cline{2-4} 
			\multicolumn{1}{c:}{Laser} & \textbf{0.83} & \textbf{0.94} & \textbf{0.95} \\
			\multicolumn{1}{c:}{Map}  & 0.73          & 0.80          & 0.87          \\
			\multicolumn{1}{l:}{Combined} & 0.80          & 0.92          & 0.92    
		\end{tabular}
		
		\subcaption{F1 Score}
		\begin{tabular}{ccccc}
			& Closed Room & Open Room & Corridor \\ \cline{2-4} 
			\multicolumn{1}{c:}{Laser} & 0.79          & \textbf{0.88} & \textbf{0.83}    \\
			\multicolumn{1}{c:}{Map}  & 0.68          & 0.80          & 0.72          \\
			\multicolumn{1}{l:}{Combined}  & \textbf{0.80} & \textbf{0.88} & 0.79        
		\end{tabular}
		
	\end{minipage}
	\begin{minipage}{0.5\textwidth}
		
		\centering
		\captionsetup[subtable]{labelformat=empty}
		\subcaption{Explored Maps}
		\vspace{-5 pt}
		\subcaption{Recall}
		
		\begin{tabular}{ccccc}
			& Closed Room              & Open Room             & Corridor                \\ \cline{2-4} 
			\multicolumn{1}{c:}{Laser} & 0.81          & 0.85         & \textbf{0.75} \\
			\multicolumn{1}{c:}{Map} & 0.72          & 0.83         & 0.69          \\
			\multicolumn{1}{l:}{Combined}  & \textbf{0.85} & \textbf{0.9} & \textbf{0.75}
		\end{tabular}
		
		\subcaption{Precision}
		\begin{tabular}{ccccc}
			& Closed Room & Open Room & Corridor \\ \cline{2-4} 
			\multicolumn{1}{c:}{Laser} & \textbf{0.87} & \textbf{0.95} & \textbf{0.95} \\
			\multicolumn{1}{c:}{Map}  & 0.82          & 0.87          & 0.98          \\
			\multicolumn{1}{l:}{Combined} & 0.86          & 0.93          & 0.93         
		\end{tabular}
		
		\subcaption{F1 Score}
		\begin{tabular}{ccccc}
			& Closed Room & Open Room & Corridor \\ \cline{2-4} 
			\multicolumn{1}{c:}{Laser}& 0.84          & 0.90          & \textbf{0.84} \\
			\multicolumn{1}{c:}{Map}  & 0.77          & 0.85          & 0.81          \\
			\multicolumn{1}{l:}{Combined}  & \textbf{0.86} & \textbf{0.91} & 0.83         
		\end{tabular}
		
	\end{minipage}
	\label{table:recall_acc_system}
	\vspace{-5pt}
\end{table*}

\par   As the robot navigates in the environment, for each update of GMapping we use three models to predict the target classes around the robot. The tracking module then uses these predictions to update the accuracy and location of the target classes on the global occupancy grid map.

\par We evaluate the final results of our tracking module by finding the closest ground-truth point to every predicted object with a maximum distance of 0.5 meters. Because we use the LiDAR and GMap local-map, details of the obstacles in each local-map improves as the robot gets closer to them.
Thus, an object can be mistakenly classified as an open-room from a distance, but as the robot comes closer, more detailed local-maps can lead to a correct prediction of an open-room (See Figure \ref{fig:closetoopen}).

\par As mentioned in Section \ref{subsection:tracking_system}, the tracking module updates the probability of target classes using all the previous model's predictions in that region. This module keeps updating the predictions until the object leaves the robot's field of view. Only then do we evaluate the predicted location.

\par If the robot tries to follow the same trajectory multiple times, the constructed local-maps (either the GMap or laser local-map) might be slightly different for the same region due to small errors and delays in navigation and performing SLAM. To address this difference, for each setting (the explored or unexplored), we repeat the experiment 30 times and report the averaged results.

\par Some examples of our system predictions for a few maps are shown in Figures \ref{fig:fr} and \ref{fig:SAIC}.
In these figures, we also annotated the false-positives and false-negative predictions of our system.

\par The recall, precision and F1 score for each target class for unexplored and explored maps are shown in Table \ref{table:recall_acc_system}. In all three types of input data, the explored maps show an increased F1 score compared to unexplored ones. Part of this increase is due to using a global frame as the origin of our tracking module. The tracking module uses the global GMap occupancy grid maps to do the frame-to-frame data association and subsequently improves the detection accuracy of the system. As a result, if we have a more stable GMap, the accuracy of tracking increases (the tracking module is provided with a more stable global map). In explored maps,  the F1 score increase is more pronounced in the Map model, which is due to a more detailed GMap local-map. Figure \ref{fig:laser_map} depicts a few examples comparing the laser and GMap local-map in the explored and unknown maps.

\subsection{Experiment Three}
\begin{figure}[t]
	\centering
	\begin{subfigure}{\columnwidth}
		\includegraphics[width=1\textwidth]{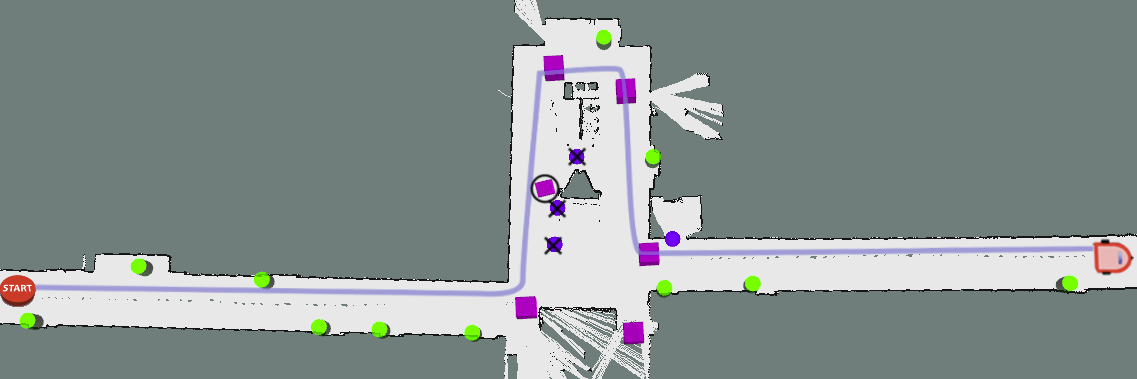}
		\caption{First part of the experiment}
	\end{subfigure}
	\hskip2em
	\begin{subfigure}{\columnwidth}
		\includegraphics[width=1\textwidth]{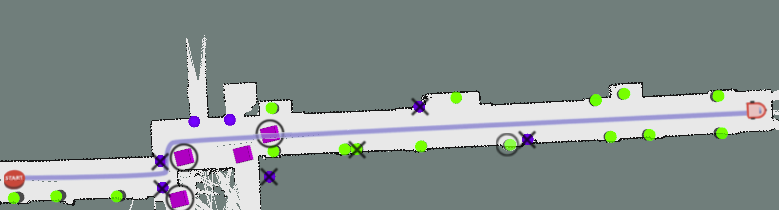}
		\caption{Second part of the experiment}
	\end{subfigure}
	
	\caption{Prediction of our system for the real world (TASC-1 building) experiment using the \textbf{Laser model}. Closed-rooms predictions are annotated by green cylinders, open-rooms by blue spheres and the start of each corridor by purple cubes. The false-positives are annotated by $\times$ and false-negatives by $\circ$ over predictions. }
	\label{fig:huskylaser}
	\vspace{-10pt}
\end{figure}

Experiment Three evaluates our system in the real world. In this experiment, we use a ClearPath Husky robot equipped with a Sick LMS-111 LiDAR and an IMU sensor. The experiment is done in two corridors of the TASC1 building at Simon Fraser University. We moved the robot in these two corridors using a joystick while recording the robot's sensor data. These corridors include a total of 26 closed-rooms, ten corridors (corridors are paths in each intersection) and three open-rooms. The trajectories and environment maps along with the predictions of the "Laser" model are shown in Figure \ref{fig:huskylaser}. 

\begin{table}[]
	\captionsetup[subtable]{labelformat=empty}
	\caption{Recall, precision and F1 Score for all three models for Experiment Three. }
	\begin{adjustwidth}{-1.1in}{-1in}
		\centering
		\subcaption{Recall}
		
		\begin{tabular}{ccccc}
			& Closed Room              & Open Room             & Corridor                \\ \cline{2-4} 
			\multicolumn{1}{c:}{Laser} & \textbf{0.96}                     & \textbf{1}                     & \textbf{0.6}                     \\
			\multicolumn{1}{c:}{Map}   & 0.81                     & 0.67                  & 0.4                     \\
			\multicolumn{1}{l:}{Combined}  & 0.77        & \textbf{1}         & 0.2     
		\end{tabular}
		
		\subcaption{Precision}
		\begin{tabular}{ccccc}
			& Closed Room & Open Room & Corridor \\ \cline{2-4} 
			\multicolumn{1}{c:}{Laser} & \textbf{0.96 }      & 0.27      & \textbf{1}    \\
			\multicolumn{1}{c:}{Map}   & 0.81        & 0.11      & 0.57     \\
			\multicolumn{1}{l:}{Combined}  & 0.91        & \textbf{0.33}      & 0.67    
		\end{tabular}
		
		\subcaption{F1 Score}
		\begin{tabular}{ccccc}
			& Closed Room & Open Room & Corridor \\ \cline{2-4} 
			\multicolumn{1}{c:}{Laser} & \textbf{0.96}       & 0.43      & \textbf{0.71}     \\
			\multicolumn{1}{c:}{Map}   & 0.81        & 0.19      & 0.47     \\
			\multicolumn{1}{l:}{Combined}  & 0.83        & \textbf{0.50}      & 0.31    
		\end{tabular}
		
	\end{adjustwidth}
	\label{table:exp3}
	\vspace{-14pt}
\end{table}

Table \ref{table:exp3} shows the precision, recall and F1 score of each target class using our three models. Based on these tables, the laser model gives a slightly better F1 score for closed-rooms and corridors compared to the other two models. This is probably due to the increased delays and accumulating sensor error while generating the GMap in the real world. We also observe that the models have some difficulties classifying corridors. In some cases, corridors are classified as open-rooms. Although recall for the open-room is high, precision is much lower due to many false-positive predictions from the network. These false positives may be caused by the additional error associated with real-world sensors.

\section{Conclusion and Future Work}

In this paper, we presented a system to describe the navigational cues around a robot using a combination of 2D LiDAR data and occupancy grid maps. We trained a CNN to predict the closed-rooms, open-rooms and intersections around the robot. A tracking module aggregated the predictions to locate and classify the navigational cues more accurately. We evaluated our system in different settings in both simulation and the real world. Based on our results, in simulation, using the combination of LiDAR data and occupancy grid maps can help to achieve a relatively higher score. However, in the real world experiment, the performance from only using 2d LiDAR data was higher. This may be due to the increase of delay and sensor error in the real world, which generates a less accurate GMap compared to the simulation.

\par We created our own dataset using eight occupancy grid maps. In the future, this study can be improved in terms of accuracy and robustness using a dataset that contains a variety of environments. Also, environmental feedback can be extended to include  more useful guidance. For instance, the system can be improved to detect steps or stairs and notify the user. Moreover, this work can be combined with our follow-ahead system \cite{nikdel:icra18} to improve the following behaviour. In particular, the robot can use our detection system to slow down near the intersections and watch for the user's reaction, in order to choose which path to follow.

\bibliographystyle{abbrvnat} 
\bibliography{main.bbl}

\end{document}